\pdfoutput=1

\documentclass[11pt]{article}

\usepackage{EMNLP2022}

\usepackage{times}
\usepackage{latexsym}

\usepackage[T1]{fontenc}

\usepackage[utf8]{inputenc}

\usepackage{microtype}

\usepackage{inconsolata}
\usepackage{graphicx}
\usepackage{float}
\usepackage{amsmath, bm}
\usepackage{color}
\usepackage{multirow}
\usepackage{booktabs}
\usepackage{bbding}
\usepackage{amssymb}
\usepackage{diagbox}
\newcommand\numberthis{\addtocounter{equation}{1}\tag{\theequation}}
\usepackage{tablefootnote}

\title{SpeechUT: Bridging Speech and Text with Hidden-Unit for Encoder-Decoder Based Speech-Text Pre-training}

\author{Ziqiang Zhang$^{1,}$\thanks{\ \ Work done during internship at Microsoft Research Asia.}\ , Long Zhou$^{2,}$\thanks{\ \ Corresponding author.}\ , Junyi Ao$^{3,*}$, Shujie Liu$^2$, Lirong Dai$^1$, Jinyu Li$^2$, Furu Wei$^2$\\
$^1$University of Science and Technology of China \\
$^2$Microsoft \\
$^3$The Chinese University of Hong Kong, Shenzhen \\
}

\begin{document}
\maketitle
\begin{abstract}
The rapid development of single-modal pre-training has prompted researchers to pay more attention to cross-modal pre-training methods.
In this paper, we propose a unified-modal speech-unit-text pre-training model, SpeechUT, to connect the representations of a speech encoder and a text decoder with a shared unit encoder.
Leveraging hidden-unit as an interface to align speech and text, we can decompose the speech-to-text model into a speech-to-unit model and a unit-to-text model, which can be jointly pre-trained with unpaired speech and text data respectively.
Our proposed SpeechUT is fine-tuned and evaluated on automatic speech recognition (ASR) and speech translation (ST) tasks. Experimental results show that SpeechUT gets substantial improvements over strong baselines, and achieves state-of-the-art performance on both the LibriSpeech ASR and MuST-C ST tasks.
To better understand the proposed SpeechUT, detailed analyses are conducted.
The code and pre-trained models are available at \url{ https://aka.ms/SpeechUT}.
\end{abstract}

\section{Introduction} \label{sec:intro}
Self-supervised pre-training with large-scale unlabeled data obtains remarkable progress on various downstream tasks \cite{devlin2018bert,radford2019language,dong2019unified,baevski2020wav2vec,hsu2021hubert,chen2021wavlm}.
Specifically, pre-trained models, such as BERT \cite{devlin2018bert} and GPT \cite{radford2019language}, have extensively promoted the development of natural language processing (NLP).
Researchers also develop many pre-trained speech models utilizing a mass of unlabeled audio data, e.g., wav2vec \cite{baevski2020wav2vec} and HuBERT \cite{hsu2021hubert}.
Although text and speech are two different modalities, they have a natural relationship because they can be viewed as two kinds of expressions of language.
Hence, joint pre-training of speech and text has received increasing attention from the research community in recent years \cite{ao2021speecht5,bapna2021slam,tang2022unified}.

One line of speech-text joint pre-training builds a shared encoder to learn speech and text representation jointly, such as SLAM \cite{bapna2021slam}, which needs a random initialization of the decoder parameter for fine-tuning an encoder-decoder model.
Another line of studies, e.g., SpeechT5 \cite{ao2021speecht5} and STPT \cite{tang2022unified}, directly pre-trains an encoder-decoder model on speech and text corpus to boost the performance of automatic speech recognition (ASR) and speech translation (ST), leveraging unsupervised vector quantization \cite{oord2017neural} and supervised speech-text data to encourage the alignment of speech and text respectively.
For these cross-modal speech-to-text models, a key problem is how to naturally connect the speech encoder and the text decoder.

\begin{figure}[t]
	\centering
	\includegraphics[width=0.9\linewidth]{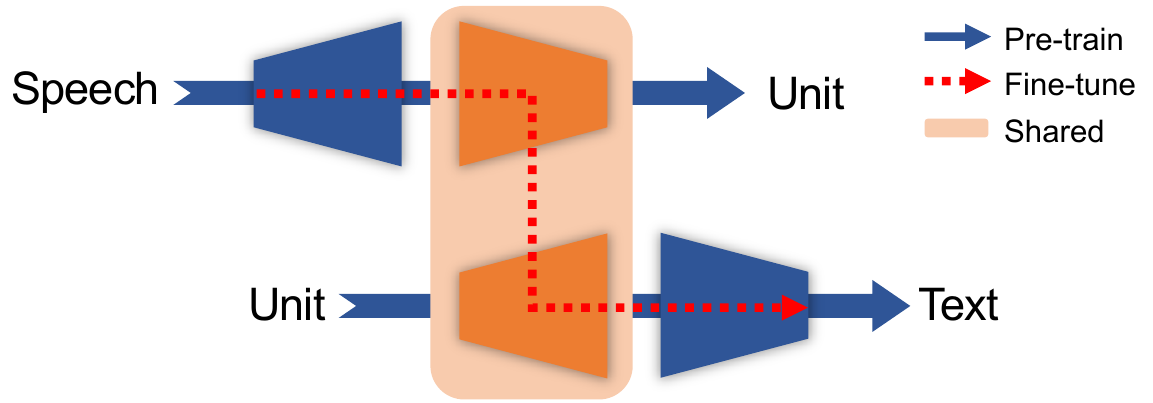}
	\caption{A high-level illustration of SpeechUT. After pre-trained with speech-to-unit and unit-to-text tasks (blue arrows), the model with a shared unit encoder enables speech-to-text tasks for fine-tuning (red arrow).} \label{fig:high_level}
\end{figure}%

Our preliminary observation shows that an intermediate hidden-unit representation \cite{hsu2021hubert} can be regarded as the bridge between speech and text modalities, and it can provide a strong mapping relationship with both of them (see Appendix \ref{Appen:cascade}).
This inspires us to leverage hidden-unit as the semantic interface between the speech encoder and the text decoder in the encoder-decoder framework, and decompose the speech-to-text model into a speech-to-unit model and a unit-to-text model, which can be pre-trained with unpaired speech and text data respectively, as shown in Figure \ref{fig:high_level}.

In this paper, we propose a unified speech-unit-text pre-training method (\textbf{SpeechUT}), using hidden-unit representation as a bridge between the speech-encoder and the text-decoder.
SpeechUT leverages three unsupervised pre-training tasks, including a speech-to-unit (S2U) task to model the mapping between speech and unit like HuBERT, masked unit modeling (MUM) task to learn better unit representation, and a unit-to-text (U2T) task to recover text from middle shared hidden-unit representation.
To generate training data for S2U, MUM, and U2T, two off-line generators trained with a small amount of paired data (100h) are introduced to produce discrete unit sequences for large-scale unpaired speech and text.
Experiments are conducted on two typical speech-to-text tasks, ASR and ST, followed by principal analysis to better understand the proposed method.
The contributions of this paper are summarized as follows,
\begin{itemize}
\item We propose a unified speech-text pre-training method SpeechUT to bridge the speech encoder and the text decoder with hidden units. 
\item We decouple the speech-to-text model into speech-to-unit and unit-to-text models, to take advantage of a large amount of unpaired speech and text data for pre-training.
\item Our proposed SpeechUT achieves state-of-the-art performance in downstream speech recognition and speech translation tasks.
\end{itemize}

\section{Related Work}
The proposed SpeechUT is built upon the Transformer encoder-decoder model \cite{vaswani2017attention} and relates to discrete speech representation learning and joint speech-text pre-training. 
We discuss these topics in the following.

\paragraph{Discrete Speech Representation Learning}
Discretizing continuous speech signals for speech representation learning has drawn substantial attention.
Vq-wav2vec \cite{baevski2019vq} and wav2vec 2.0 \cite{baevski2020wav2vec} attempt at discretizing speech signals into quantized units from a learnable codebook \cite{oord2017neural}.
PBERT \cite{wang2022supervision} instead uses phonemes as the discrete targets in a semi-supervised setting.
SemFace \cite{ren2021semface} proposes to use language-independent vector quantized units as the semantic interface of encoder pre-training and decoder pre-training.
Inspired by the masked language model in BERT \cite{devlin2018bert}, HuBERT \cite{hsu2021hubert} first introduces the masked speech prediction of hidden units to pre-train a universal speech model.
Particularly, the hidden units can be clustered from log Mel-filterbank features or the hidden states of the previous pre-trained model.
Recently, some studies explore leveraging the discrete hidden units to build speech-to-speech translation systems \cite{lee2021direct,lee2021textless}, which first convert source speech into target units, then generate the target waveform from predicted units.
However, our goal in this paper is to jointly pre-train speech and text with the hidden units as the intermediate bridge.

\begin{figure*}[t]
	\centering
	\includegraphics[width=0.9\textwidth]{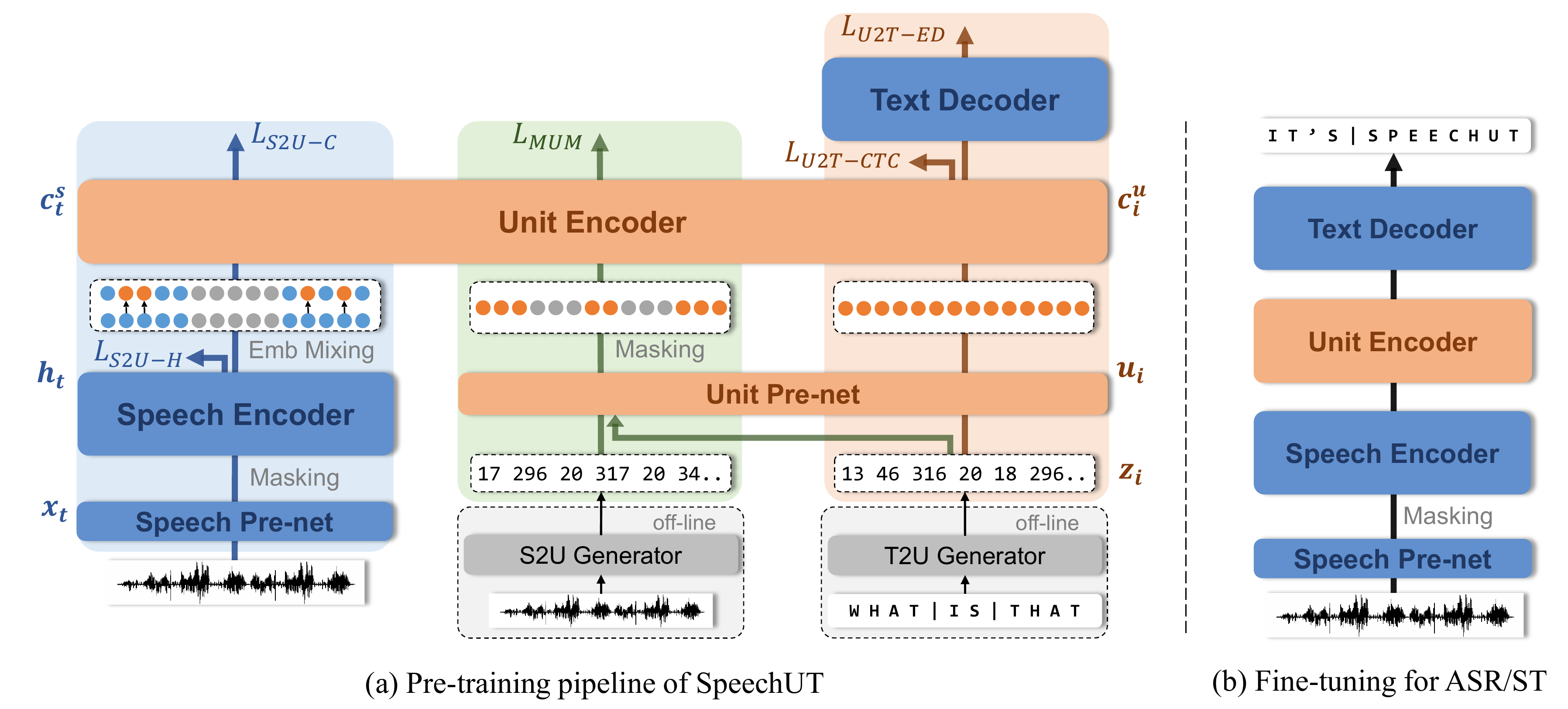}
	\caption{(a) The overall framework of SpeechUT, which is pre-trained with the speech-to-unit (\textbf{\textcolor[RGB]{47,85,151}{S2U}}) task, the masked unit modeling (\textbf{\textcolor[RGB]{56,87,35}{MUM}}) task and the unit-to-text (\textbf{\textcolor[RGB]{132,60,12}{U2T}}) task jointly. The discrete units are extracted from off-line speech-to-unit (S2U) and text-to-unit (T2U) generators. (b) Fine-tuning is performed for speech-to-text tasks by cascading the speech encoder, the unit encoder, and the text decoder into an end-to-end model.}\label{architecture}
\end{figure*}%

\paragraph{Joint Speech-Text Pre-Training}
Single-modal pre-trained models have achieved remarkable results in both natural language processing and spoken language processing, such as BERT \cite{vaswani2017attention}, UniLM \cite{dong2019unified}, XLNet \cite{yang2019xlnet}, wav2vec 2.0 \cite{baevski2020wav2vec}, HuBERT \cite{hsu2021hubert}, and WavLM \cite{chen2021wavlm}.
Thanks to the rapid development of these single-modal pre-training works, researchers begin to pre-train a cross-modal model with both speech and text data \cite{chung2020splat,kim2021st,qian2021speech,ao2021speecht5,bapna2021slam,zhang2022yitrans,tang2022unified}.
One category of these works focuses on pre-training a unified encoder model for spoken language understanding \cite{chung2020splat,kim2021st,qian2021speech,zhang2022speechlm}.
In parallel to our work, SpeechLM \cite{zhang2022speechlm} leverages two kinds of tokenizers to tokenize speech and text, and aims at unifying speech and text modalities into the same semantic space within one encoder model.
When fine-tuning an encoder-decoder model, a randomly initialized decoder needs to be superimposed on the encoder for speech-to-text tasks \cite{bapna2021slam,bapna2022mslam}.
Besides, Maestro \citep{chen2022maestro} utilizes paired speech-text data to learn speech-text alignment through a modality-matching algorithm in RNN-T framework.
Our proposed SpeechUT model is most related to encoder-decoder pre-trained models like SpeechT5 \cite{ao2021speecht5} and STPT \cite{tang2022unified}, in which speech and text are directly connected by a shared encoder.
Unlike them, SpeechUT leverages hidden units \cite{hsu2021hubert} as the bridge between the speech encoder and the text decoder, decoupling the conventional model into two pre-trained speech-to-unit and unit-to-text models.

\section{SpeechUT}
Figure \ref{architecture} shows the overall framework of SpeechUT, which leverages the unit representation as the bridge between speech and text.
In this section, we will introduce the model architecture, pre-training, and fine-tuning methods.

\subsection{Model Architecture}
As illustrated in Figure \ref{architecture}(a), SpeechUT mainly contains a speech encoder, a unit encoder, and a text decoder.
In addition, speech and unit pre-nets pre-process the input waveform and the text tokens into fixed-dimensional hidden states, respectively.

\paragraph{Speech/Unit Pre-nets}
The speech pre-net is a stack of 1-D convolutional layers with 512 channels and kernel sizes of [10,3,3,3,3,2,2].
The overall downsampling rate is 320.
Given a 16K Hz speech waveform, the speech pre-net will convert it into a sequence of speech features, $\bm{X}=(x_1, x_2, \dots, x_{T})$, where $T$ is the sequence length.
The unit pre-net is a simple embedding layer which converts a sequence of unit tokens, $\bm{Z}=(z_1, z_2, \dots, z_{L})$, into latent embeddings, $\bm{U}=(u_1, u_2, \dots, u_{L})$, where $L$ is the sequence length.
The latent embeddings are then equipped with learned positional encodings.

\paragraph{Speech Encoder}
The speech encoder is a stack of Transformer layers \cite{vaswani2017attention} that transforms the local speech features $\bm{X}$ into contextualized speech hidden states, $\bm{H}=(h_1, h_2, \dots, h_{T})$.

\paragraph{Unit Encoder}
The unit encoder has the same architecture and layer numbers as the speech encoder.
It is designed to align the speech hidden states $\bm{H}$ and the unit embeddings $\bm{U}$ into the same latent space.
The unit encoder takes two types of input, $\bm{H}$ and $\bm{U}$, and outputs high-level contextualized representations, $\bm{C^s}=(c^s_1, c^s_2, \dots, c^s_{T})$, and $\bm{C^u}=(c^u_1, c^u_2, \dots, c^u_{L})$, respectively.

\paragraph{Text Decoder}
The text decoder is a Transformer decoder \cite{vaswani2017attention} consisting of a text embedding layer, stacked Transformer layers, and a text output layer.
It is used to generate the target text sequence $\bm{Y}=({y}_1,{y}_2,\dots, {y}_{|\bm{Y}|})$ from left to right according to the output of the unit encoder.

\subsection{Pre-Training Tasks}
To pre-train the components of SpeechUT, we propose three pre-training tasks:

\paragraph{Speech-to-Unit (S2U) Task}
The speech-to-unit objective is similar to HuBERT \cite{hsu2021hubert}, where the model needs to predict the unit of the masked positions based on the non-mask regions in a speech sequence.
Particularly, SpeechUT enables this prediction for both the output of the speech encoder ($\bm{H}$) and the output of the unit encoder ($\bm{C^s}$),
\begin{equation}
\begin{aligned}
  & \mathcal{L}_{S2U}  = \mathcal{L}_{S2U-H} + \mathcal{L}_{S2U-C} \\
  & = - \sum_{t\in \mathcal{M}}\left(\text{log}{~p\left(z_t|h_t\right)} + \text{log}{~p\left(z_t|c^s_t\right)}\right)
    \label{eq:L_S2U}
\end{aligned}
\end{equation}
where $\mathcal{M}$ is a set of masked positions and $z_t$ is the corresponding unit at position $t$.
$p(.)$ computes the probabilities, i.e.,
\begin{align*} 
    p(z|h_t) &= \frac{\exp(\text{cos}(\bm{W}^{s}h_t, \bm{e}_z)/\tau)}{\sum_{z^{'}\in \mathcal{Z}} \exp(\text{cos}(\bm{W}^{s}h_t, \bm{e}_{z^{'}})/\tau)} \numberthis \\
    p(z|c^s_t) &= \text{softmax}\left(\bm{W}^{u}c^s_t\right) \numberthis \label{eq:speech_proj}
\end{align*}
where $\bm{W}^{s}$ and $\bm{W}^{u}$ are projection weights, $\tau$ is the temperature coefficient set to 0.1,  and $\mathcal{Z}$ is the set of unit categories.
$\text{cos}(.)$ computes cosine similarity between two vectors following HuBERT \cite{hsu2021hubert}.
Here $\bm{e}$ is a unit embedding matrix preserved by the speech encoder, it does not share parameters with the unit pre-net since HuBERT uses a lower embedding dimension.

\paragraph{Unit-to-Text (U2T) Task}
SpeechUT performs the unit-to-text task as a regular encoder-decoder based sequence-to-sequence task \cite{vaswani2017attention}.
The text sequence serves as the target and the corresponding generated unit sequence serves as the input.
Conditioned on the output of the unit encoder, $\bm{C^u}$, the loss is formulated as
\begin{equation}
    \mathcal{L}_{U2T-CE} = - \sum_{i=1}^{|\bm{Y}|} \text{log}~p({y}_i|\bm{Y}_{<i}, \bm{C^u})
\end{equation}
where $\bm{Y}=({y}_1,{y}_2,\dots, {y}_{|\bm{Y}|})$ is the text sequence and $\bm{Y}_{<i}$ is its prefix from position $0$ to position $i$.
$p(.)$ is parameterized by a linear softmax layer.

Besides, to enhance the unit-to-text generation, following \cite{shinji2017hybrid} we formulate a joint CTC \cite{Graves10.1145CTC} objective which directly predicts the target text sequence from the unit encoder,
\begin{align*}
    &\mathcal{L}_{U2T-CTC} = - \text{log}~p_{CTC}(\bm{Y} | \bm{C^u}) \numberthis \\
    &\mathcal{L}_{U2T} = \mathcal{L}_{U2T-CE} + \mathcal{L}_{U2T-CTC} \numberthis
\end{align*}
where $p_{CTC}(.)$ is parameterized by a single 1-D convolutional layer with a kernel size of 2 and channel of 768, followed by a linear projection to the text vocabulary.

\paragraph{Masked Unit Modeling (MUM) Task}
Note that in S2U and U2T tasks, the unit serves as the target and the input, respectively.
To enhance the unit-in, unit-out property, inspired by BERT \cite{vaswani2017attention} and HuBERT \cite{hsu2021hubert}, SpeechUT performs an additional masked unit modeling (MUM) task, with the training data combining all the units in S2U and U2T tasks.
The unit encoder needs to predict the unit categories of the masked positions in a unit sequence, with loss formulated as
\begin{equation}
    \mathcal{L}_{MUM} = - \sum_{i\in \mathcal{M}}\text{log}{~p(z_i|c^u_i)}
\end{equation}
where $\mathcal{M}$ is a set of masked positions and the probability $p(.)$ is computed as 
\begin{equation}
    p(z|c^u_i) = \text{softmax}\left(\bm{W}^{u}c^u_i\right) \label{eq:unit_proj}
\end{equation}

\paragraph{Multi-task Learning}
In the pre-training stage, SpeechUT performs multi-task pre-training with three tasks,
\begin{equation}
    \mathcal{L} = \mathcal{L}_{S2U} + \lambda \mathcal{L}_{U2T} + \gamma \mathcal{L}_{MUM}
\end{equation}
where $\lambda$ and $\gamma$ control the balance of losses.
During multi-task learning, SpeechUT is expected to connect the speech encoder and the text decoder by the unit encoder.
Thus the data could flow smoothly from the speech input end to the text output end even without consuming speech-text paired data.

\subsection{Hidden-Unit Generation} \label{ssec:unit-generator}
Using these three tasks for pre-training, we need to construct three kinds of training data, the unit data, the speech-unit paired data, and the unit-text paired data. The unit data is the combination of the units in speech-unit and unit-text data.
To get the latter two, we introduce two off-line unit generators, the speech-to-unit (S2U) generator and the text-to-unit (T2U) generator.
The S2U generator could be any off-line unsupervised clustering model that discretizes the unlabeled speech sequences into the hidden units, e.g., the k-means model learned from HuBERT \cite{hsu2021hubert}.
Besides, our T2U generator is a sequence-to-sequence model \cite{vaswani2017attention}.
As the units generated from text should have the same style as the units generated from speech, we leverage a small amount of paired ASR data{\footnote{A small amount of ASR data is enough to train the T2U generator (see Appendix \ref{Appen:cascade}).}} to train the T2U generator.
Specifically, we generate the units from speech for a small paired dataset using the S2U generator, and then remove the repetitive units of adjacent frames to get \textit{reduced} units \cite{ao2022pre}. The \textit{reduced} units and the corresponding transcription form the training data for the T2U generator.
With the trained T2U generator, large-scale unpaired text corpora can be converted to a large unit-text paired corpus for the U2T pre-training task.


\subsection{Embedding Mixing Mechanism}
Multi-task learning assumes the representations of different modalities are aligned to the same latent space \cite{Wei9747760Optimizing}.
However, we found that the unit encoder always performs two individual tasks for speech and unit without providing explicit alignment information between them.
To better align the speech and unit representations in the unit encoder, in this paper we adopt a simple embedding mixing mechanism for S2U task, which is to mix the embeddings of two modalities in one sequence.
Since each unlabeled speech sequence has the generated units at each time position, we randomly replace a portion of speech hidden states $h_t$ in the sequence with the corresponding unit embeddings $u_t$, i.e.,
\begin{equation}
    h_t^{'} = 
    \begin{cases}
    u_t & t \in \mathcal{R} - \mathcal{M} \\
    h_t & \text{otherwise}
    \end{cases}
\end{equation}
where $\mathcal{M}$ is the masked positions in Eqn. (\ref{eq:L_S2U}) and $\mathcal{R}$ is a set of randomly selected positions.
$\mathcal{R} - \mathcal{M}$ means the embedding mixing is restricted by only operating on the non-mask positions.
Different from previous work \cite{chen2022maestro,Wei9747760Optimizing,fang-etal-2022-stemm}, which rely on force-aligned phoneme or word labels, SpeechUT uses units for mixing, making it available for full of unlabeled data.

\subsection{Fine-Tuning for ASR and ST}
After pre-training, we drop the unit pre-net and stack the speech encoder, the unit encoder and the text decoder into a complete sequence-to-sequence model, which can be fine-tuned for any speech-to-text task, such as ASR and ST.
Note that all modules have been pre-trained, including the text output layer, and no new parameters are introduced in the fine-tuning stage.

\section{Experiments}

\begin{table*}[t!]
\begin{center}
\resizebox{\linewidth}{!}{
\begin{tabular}{lc|ccc|cc|ccc}
\toprule
\multirow{2}{*}{Model}                 & \multirow{2}{*}{Size}  & \multicolumn{3}{c|}{Pre-training Data}        & \multicolumn{2}{c|}{WER ($\downarrow$) Without LM} & \multicolumn{3}{c}{WER ($\downarrow$) With LM}   \\
                                       &                        & Speech           & Paired         & Text      & test-clean          & test-other      & LM        & test-clean& test-other \\
\midrule
\multicolumn{7}{l}{\textit{960h hours pre-trained}}   \\
\midrule
wav2vec 2.0 \cite{baevski2020wav2vec}  & Base (0.1B)            & 960h             & -               & -         & 6.1                & 13.3            & 4-gram    & 3.4       & 8.0 \\
HuBERT \cite{hsu2021hubert}            & Base (0.1B)            & 960h             & -               & -         & 6.3                & 13.2            & 4-gram    & 3.4       & 8.1 \\
WavLM \cite{chen2021wavlm}             & Base (0.1B)            & 960h             & -               & -         & 5.7                & 12.0            & 4-gram    & 3.4       & 7.7 \\
ILS-SSL \cite{Wang9747022ILS}          & Base (0.1B)            & 960h             & -               & -         & 4.7                & 10.1            & 4-gram    & 3.0       & 6.9 \\
data2vec \cite{baevski2022data2vec}    & Base (0.1B)            & 960h             & -               & -         & 4.2$^\ast$         & 9.7$^\ast$      & 4-gram    & 2.8       & 6.8 \\
PBERT \cite{wang2022supervision}       & Base (0.15B)           & 960h             & 100h$^\dagger$  & -         & 4.7                & 10.7            & 4-gram    & 3.1       & 7.3 \\ 
SpeechT5 \cite{ao2021speecht5}         & Base (0.15B)           & 960h             & -               & 40M       & 4.4                & 10.4            & Transf.   & 2.4       & 5.8 \\
Speech2C \cite{ao2022pre}              & Base (0.15B)           & 960h             & -               & -         & 4.3                & 9.0             & Transf.   & 2.4       & 5.2 \\
Wav2seq \cite{wu2022wav2seq}           & Base (0.15B)           & 960h             & -               & -         & -                  & 11.2            & -         & -         & - \\      
wav2vec 2.0 \cite{baevski2020wav2vec}  & \textbf{Large} (0.3B)  & 960h             & -               & -         & 4.7                & 9.0             & Transf.   & 2.3       & 5.0 \\
Baseline (Ours)                        & Base (0.15B)           & 960h             & -               & 40M       & 3.8                & 8.0             & Transf.   & 2.3       & 5.1 \\
\textbf{SpeechUT (Ours)}               & Base (0.15B)           & 960h             & 100h$^\dagger$  & 40M       & \textbf{2.7}       & \textbf{6.8}    & Transf.   & \textbf{2.0}  & \textbf{4.5} \\
\midrule
\multicolumn{7}{l}{\textit{60kh hours pre-trained}}     \\
\midrule
wav2vec 2.0 \cite{baevski2020wav2vec}  & \textbf{Large} (0.3B)  & \textbf{60k}h    & -               & -         & 3.1                & 6.3             & Transf.   & 2.0       & 4.0 \\
HuBERT \cite{hsu2021hubert}            & \textbf{Large} (0.3B)  & \textbf{60k}h    & -               & -         & -                  & -               & Transf.   & 2.1       & 3.9 \\
WavLM \cite{chen2021wavlm}             & \textbf{Large} (0.3B)  & \textbf{94k}h    & -               & -         & -                  & -               & Transf.   & 2.1       & 4.0 \\
ILS-SSL \cite{Wang9747022ILS}          & \textbf{Large} (0.3B)  & \textbf{60k}h    & -               & -         & 2.9                & 5.8             & Transf.   & 2.0       & 4.0 \\
STPT \cite{tang2022unified}            & Base (0.16B)           & \textbf{60k}h    & 100h            & 40M       & 3.5                & 7.2             & -         & -         & - \\
\textbf{SpeechUT (Ours)}               & \textbf{Large} (0.38B) & \textbf{60k}h    & 100h$^\dagger$  & 40M       & \textbf{2.2}       & \textbf{4.5}    & Transf.   & \textbf{1.9}  & \textbf{3.6} \\
\bottomrule
\end{tabular}
}
\end{center}
\caption{\label{Tab: wer_100h} ASR performance on 100-hour LibriSpeech benchmark. Speech/Paired/Text indicates the unlabeled speech data, the paired ASR data, and the unpaired text data respectively. $^\ast$ indicates our reproduction results, and $^\dagger$ indicates the data is not directly used for pre-training.}
\end{table*}

\subsection{Dataset}
We conducted experiments individually for the ASR task on English and ST tasks in three directions: English (En) to German (De), Spanish (Es), and French (Fr).
For ASR pre-training, the S2U task uses unlabeled speech data from LibriSpeech \cite{panayotov2015LibriSpeech} and  LibriLight \cite{kahn2020libri}, which contain about 960 and 60,000 hours of speech respectively.
U2T task uses text from LibriSpeech LM Corpus\footnote{http://www.openslr.org/11/}, containing about 40M sentences.
MUM task uses the combination of units generated from the speech and the text.

For ST pre-training, the S2U task uses unlabeled speech data from LibriSpeech and MuST-C \cite{di-gangi-etal-2019-must}. The latter contains hundreds of hours of speech (see Appendix \ref{Appen:data}).
U2T task only optimizes $\mathcal{L}_{U2T-ED}$ and uses the paired machine translation (MT) data from WMT datasets, where the English-side text is used to generate units, and the target-side text serves as the target of the text decoder.
WMT contains about 4.6M, 15M and 40M paired sentences for En-\{De\footnote{https://www.statmt.org/wmt16/},Es\footnote{https://www.statmt.org/wmt13/},Fr\footnote{https://www.statmt.org/wmt14/}\}, respectively.
MUM task also uses the combination of units from two sources.

The T2U generator is trained on LibriSpeech 100 hours subset (\texttt{train-clean-100}) and used for both ASR and ST pre-training.
For downstream tasks, we use LibriSpeech 100 and 960 hours training set for ASR fine-tuning and MuST-C En-De/Es/Fr train sets for ST fine-tuning.
More details about the dataset and the text pre-processing can be found in Appendix \ref{Appen:data}.

\subsection{Model Configuration}
\paragraph{SpeechUT}
The base model consists of 6 Transformer layers with relative positional attention bias \cite{shaw-etal-2018-self} for all encoder/decoders.
The model dimension is 768 and the FFN dimension is 3072.
The large model scales up to 12 Transformer layers for the speech/unit encoder with the model dimension of 1024 and the FFN dimension of 4096, and 12 Transformer layers for the text decoder without changing model dimensions.
We use the character vocabulary for ASR tasks and 10k SentencePiece \cite{kudo-richardson-2018-sentencepiece} for ST tasks.
CTC prediction head is not applied for ST tasks.
The total parameter size is about 156M for the ASR base model, 162M for the ST base model, and 380M for ASR large model.


\paragraph{Baseline} \label{ssec: baselines}
For comparison, we also implement a baseline with similar architecture but without using units as an intermediate modality.
The baseline combines the Speech2C \cite{ao2022pre} task with the BART \cite{lewis2019bart} task to perform multi-task pre-training.
Specifically, Speech2C takes speech as input and predicts the corresponding units at the decoder.
BART takes the corrupted character-level text sequence as input and predicts the complete sequence at the decoder.
The baseline consists of a shared 12-layer encoder and a shared 6-layer decoder.
The model size is the same as the SpeechUT base model.

\paragraph{Unit Generators} \label{ssec: Generators}
The S2U generator is the k-means model learned from the released HuBERT base model \cite{hsu2021hubert}, which has 500 unit classes.
The T2U generator has 6 Transformer layers for both the encoder and the decoder, with a model dimension of 768 and the FFN dimension of 3072.

\subsection{Training Details}
All the experiments are conducted in Fairseq \cite{ott2019fairseq}.
The loss weights $(\lambda, \gamma)$ are set to $(0.1, 0.5)$ for ASR pre-training and $(1.0, 0.5)$ for ST pre-training.
Before each optimization step, the model simultaneously consumes batched data from 3 tasks and accumulates their gradients.
The masking in S2U and MUM tasks follows the same configuration with HuBERT \cite{hsu2021hubert}, with the mask probability of 8\% and the mask length of 10.
The selection of $\mathcal{R}$ also follows the masking strategy, but with the probability of 4\% and the window length of 5.

For ASR fine-tuning, we keep the pre-trained CTC prediction head and tune a CTC/Attention multi-task model \cite{shinji2017hybrid} with the CTC weight of 0.5.
For ST fine-tuning, only encoder-decoder loss is optimized.
More details about pre-training and fine-tuning can be found in Appendix \ref{Appen:training}.

\begin{table*}[ht]
\begin{center}
\renewcommand\arraystretch{1.0}
\small
\begin{tabular}{lc|ccc|ccc}
\toprule
\multirow{2}{*}{Models}     & \multirow{2}{*}{Sizes}    & \multicolumn{3}{c|}{Pre-training Data}                    & \multicolumn{3}{c}{Fine-tuning BLEU ($\uparrow$)} \\
                                    &                   & Speech (h)        & ASR (h)           & MT (\#utt)        & En-De             & {En-Es}           & {En-Fr} \\
\midrule
FAT-ST \cite{pmlr-v139-zheng21a}    & -                 & 3.7k              & 1.4$\sim$1.5k     & 1.9$\sim$2.0M     & 25.5              & 30.8              & -      \\
SATE \cite{xu-etal-2021-stacked}    & -                 & -                 & 1.4k              & 18M               & 28.1              & -                 & -      \\
STEMM \cite{fang-etal-2022-stemm}   & -                 & 960               & 408$\sim$504      & 4.6$\sim$40M      & 28.7              & 31.0              & 37.4  \\
ConST \cite{ye2022cross}            & 0.15B             & 960               & 408$\sim$504      & 4.6$\sim$40M      & 28.3              & 32.0              & 38.3   \\
STPT \cite{tang2022unified}         & 0.16B             & 60k               & 408$\sim$504      & 4.6$\sim$40M      & 29.2\tablefootnote{En-De result is from their released code.}  & 33.1              & 39.7    \\
\textbf{SpeechUT} (Ours)            & 0.16B             & 1.4$\sim$1.5k     & 100$^\dagger$       & 4.6$\sim$40M      & \textbf{30.1}     & \textbf{33.6}     & \textbf{41.4}\\
\bottomrule
\end{tabular}
\end{center}
\caption{\label{Tab: bleu} ST performance on MuST-C dataset. Speech/ASR/MT indicates auxiliary unlabeled speech data, ASR data, and MT data. $^\dagger$ indicates the data is not directly used for pre-training.}
\end{table*}

\subsection{Evaluation on Speech Recognition}
We evaluate the performance by the word error rate (WER) computed on LibriSpeech \texttt{test-clean} and \texttt{test-other} sets.
We also leverage an external Transformer language model (Transf. LM) for shallow fusion \cite{gulcehre2015using}.
The LM has a similar size to that used in the previous works and is trained on LibriSpeech LM Corpus (see Appendix \ref{Appen:inference}).
The results are summarized in Table \ref{Tab: wer_100h}, compared with several previous self-supervised approaches, 
including encoder-based models like wav2vec 2.0 \cite{baevski2020wav2vec}, HuBERT \cite{hsu2021hubert}, data2vec \cite{baevski2022data2vec}, and PBERT \cite{wang2022supervision}, and encoder-decoder models like SpeechT5 \cite{ao2021speecht5} and STPT \cite{tang2022unified}.


Table \ref{Tab: wer_100h} shows that SpeechUT outperforms all the encoder-based models by a large margin.
Our base model even behaves better than the large model of wav2vec 2.0 with 960 hours of pre-training data.
SpeechUT also outperforms all the previous encoder-decoder speech-text pre-trained models, including SpeechT5, STPT, and our baseline, achieving a new state-of-the-art performance on the \texttt{train-clean-100} set. 
Moreover, the SpeechUT Large with LM gets the WER of 1.9 and 3.6 on \texttt{test-clean} and \texttt{test-other} sets.
Due to the space limitation, the results using 960 hours of training data are given in Appendix \ref{Appen:960}.


\subsection{Evaluation on Speech Translation}
We evaluate the proposed SpeechUT on En-\{De, Es, Fr\} language pairs.
The results are shown in Table \ref{Tab: bleu}, with a comparison to recent state-of-the-art approaches, such as ConST \cite{ye2022cross} and STPT \cite{tang2022unified}.
For convenience, we reuse the off-line T2U generator trained on LibriSpeech, which is inevitably related to external 100-hour ASR data.
But we do not use any ASR labels of MuST-C as all the previous works do, which is much more than 100 hours.
As shown in Table \ref{Tab: bleu}, our SpeechUT achieves the performance of 30.1, 33.6, and 41.4 BLEU scores on En-De, En-Es, and En-Fr, respectively, demonstrating the superiority of SpeechUT over previous works. 
Specifically, SpeechUT outperforms the previous state-of-the-art methods by at most +1.7 BLEU (En-Fr) with significantly less pre-training data.

\section{Analysis \& Discussion}

\subsection{Ablation Study}
To better understand the effect of each component of SpeechUT, we pre-train different models in the absence of different tasks as well as the embedding mixing mechanism.
Specifically, these models are pre-trained on 960 hours of speech and fine-tuned on \texttt{train-clean-100}.
The results are listed in Table \ref{Tab: ablation}.
First, the embedding mixing mechanism has the biggest impact, as the absence leads to the biggest degeneration of WER, which demonstrates its importance and effectiveness.
Second, it can be noticed that the CTC loss, as a part of the U2T task, has a minor influence (0.1\textasciitilde0.2 WER) on the fine-tuning performance.
Finally, while the MUM loss has the minimum effect, we speculate that the U2T task has already modeled the unit well.

\begin{table}[t]
\begin{center}
\small
\begin{tabular}{ccc|c|cc}
\toprule
$\mathcal{L}_{S2U}$ & $\mathcal{L}_{U2T}$           & $\mathcal{L}_{MUM}$ & Mix    & dev-c & dev-o \\
\midrule
\checkmark  & \checkmark                            & \checkmark    & \checkmark    & 2.5   & 7.0        \\
\checkmark  & \footnotesize{$w\backslash o$ CTC}    & \checkmark    & \checkmark    & 2.6   & 7.2        \\
\checkmark  & \footnotesize{$w\backslash o$ CTC}    & \textbf{--}   & \checkmark    & 2.7   & 7.3        \\
\checkmark  & \footnotesize{$w\backslash o$ CTC}    & \textbf{--}   & \textbf{--}   & 3.0   & 7.9        \\
\bottomrule
\end{tabular}
\end{center}
\caption{\label{Tab: ablation} Ablation study. The performance is evaluated by WER on \texttt{dev-clean} and \texttt{dev-other} set after fine-tuning on \texttt{train-clean-100} set.}
\end{table}

\begin{table}[htb]
\begin{center}
\small
\resizebox{\linewidth}{!}{
\begin{tabular}{p{3.8cm}p{0.7cm}<{\centering}|p{0.3cm}<{\centering}p{0.3cm}<{\centering}p{0.3cm}<{\centering}p{0.3cm}<{\centering}}
\toprule
\multirow{2}{*}{Model}                                      & \multirow{2}{*}{Size}     & \multicolumn{2}{c}{Dev}   & \multicolumn{2}{c}{Test}      \\
                                                            &                           & clean     & other         & clean     & other             \\
\midrule
\emph{Self-training} \cite{Xu9414641Self}                   & 300M                      & 2.2       & 4.6           & 2.4       & 5.0               \\
\emph{Semi-supervised pre-training}                         & 156M                      & 2.4       & 4.9           & 2.5       & 5.1                \\
\midrule
SpeechUT                                                    & 156M                      & \textbf{1.6} & 4.5        & 2.0       & 4.5               \\
SpeechUT + \emph{Self-training}                                    & 156M                      & 1.7       & \textbf{4.0}  & \textbf{1.9} & \textbf{4.2}   \\
\hline
\end{tabular}
}
\end{center}
\caption{\label{Tab: self-training} ASR performance (WER) using LibriSpeech 100-hour supervised and 860-hour unsupervised speech data. LM is used for decoding.}
\end{table}

\begin{figure}[t]
	\centering
	\includegraphics[width=\linewidth]{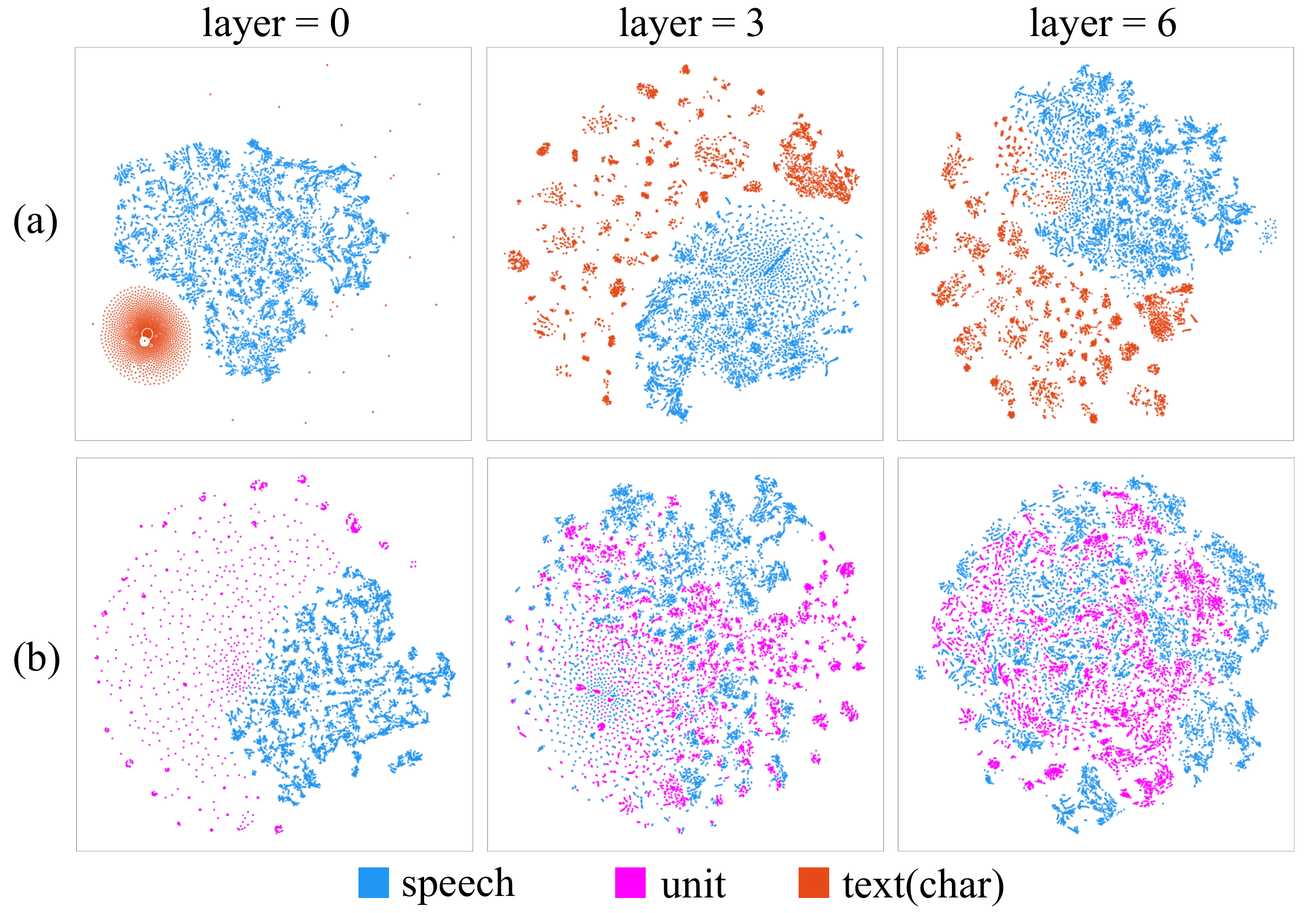}
	\caption{2-D illustration of the token-level representations of the unit encoder. (a) Semi-supervised pre-trained model; (b) SpeechUT.}\label{fig: tsne}
\end{figure}%

\subsection{Effect of Paired Data Usage} \label{ssec:semi}
SpeechUT employs a small amount of paired ASR data to train the T2U generator.
Here, we analyze and verify our method on using the ASR data compared with the other two methods, including 1) self-training \cite{Xu9414641Self} and 2) semi-supervised pre-training which combines supervised and unsupervised learning in a multi-task pre-training process like mSLAM \cite{bapna2022mslam}. Since the ASR result in this setting is not reported in mSLAM, we implemented the semi-supervised pre-training based on encoder-decoder model\footnote{Specifically, it combines the speech-to-unit task (Eqn. (\ref{eq:L_S2U})), the text-to-text BART \cite{lewis2019bart} task and the supervised CTC/Attention ASR task jointly for pre-training.}.
Experimental results in Table \ref{Tab: self-training} show our model behaves better than the self-training and semi-supervised pre-training.
Moreover, our method has the following advantages: (1) instead of using pseudo text in self-training, SpeechUT uses real text data for the decoder pre-training;
(2) simply semi-supervised pre-training only learns speech-text alignment within a small amount of paired data, while SpeechUT could align large-scale unpaired speech and text with units as a bridge;
(3) SpeechUT is also complementary to self-training, achieving further performance improvement, as shown in Table \ref{Tab: self-training}.

\subsection{Is the Encoder Getting Better?}
The U2T and MUM tasks can pre-train the unit encoder with the generated unit and unit-text data. 
Here, we attempt at evaluating the effect of U2T and MUM losses on the encoder and judging whether they can help the encoder learn better.
We first fine-tune a CTC model based on SpeechUT encoder, which obtains the 3.8 and 9.7 WER on \texttt{dev-clean} and \texttt{test-other} sets as shown in Table \ref{Tab: enc-only}.
Second, we pre-train an encoder without the text decoder and U2T task, and pre-train another encoder model by further removing the MUM task, whose results are summarized in the last two lines of Table \ref{Tab: enc-only}.
The evaluation demonstrates that our joint speech-unit-text pre-training method can still boost the performance of the encoder-only model, which means the encoder itself also learns better with U2T and MUM tasks.

\begin{table}[ht]
\begin{center}
\small
\begin{tabular}{l|cc}
\toprule
Pre-trained model                                   & dev-clean         & dev-other                       \\
\midrule
SpeechUT (\footnotesize{$w\backslash o$ decoder})   & 3.8           & 9.7                         \\
\midrule
- $w\backslash o$ $\mathcal{L}_{U2T}$                       & 4.3           & 10.3                        \\
- $w\backslash o$ $\mathcal{L}_{U2T}$, $\mathcal{L}_{MUM}$  & 4.5           & 10.7                        \\
\bottomrule
\end{tabular}
\end{center}
\caption{\label{Tab: enc-only} ASR performance (WER) of encoder-only CTC models. SpeechUT ({\footnotesize{$w\backslash o$ decoder}}) is fine-tuned by discarding the pre-trained decoder, other models are pre-trained \& fine-tuned without decoders.}
\end{table}

\begin{table}[h]
\begin{center}
\small
\begin{tabular}{c|ccc}
\toprule
Total         & Vowels        & Consonants    & Silence  \\
\midrule
85.4\%        & 79.6\%        & 85.5\%        & 96.7\%         \\
\bottomrule
\end{tabular}

\end{center}
\caption{\label{Tab: phone_align} Proportion where the paired speech and unit representations agree to the same phonemes.}
\end{table}

\subsection{Are the Speech and the Unit Aligned?}
SpeechUT aims to align the representations of speech and unit using the unit encoder as a bridge, so that information can flow smoothly from the speech end to the text end.
To verify this, we first demonstrate the alignment by validating the data distribution, i.e., the speech representation and the unit representation should follow the same distribution if they are aligned.
Figure \ref{fig: tsne} plots the token-level representations of different layers of the unit encoder (layer=0 indices the inputs).
The data are sampled from unpaired speech and unit sequences from LibriSpeech \texttt{dev-clean} set.
T-SNE \cite{van2008visualizing} is performed to reduce the dimension to 2D.
Figure \ref{fig: tsne}(a) shows that the representations are divided into two distinct regions for speech and text respectively in the semi-supervised pre-trained model, which means the model processes the two kinds of inputs independently, leading to no alignment between speech and text.
While, SpeechUT shows another behavior as shown in Figure \ref{fig: tsne}(b), where the hidden states of the two modalities are mapped to the same distribution as the layer increases.

We further validate the alignment by a linear phoneme classifier.
Specifically, we train the linear phoneme classifier using fixed speech representations extracted from the 6-th layer of the unit encoder paired with frame-level phoneme labels\footnote{ \url{http://www.kaldi-asr.org/downloads/build/6/trunk/egs/librispeech/}. The phoneme inventory size is 42, including 15 vowels, 24 consonants, and 3 silence phones.} on \texttt{train-clean-100} set.
The classifier is then tested by the unit inputs on \texttt{dev-clean} set to see whether they predict the same phonemes with their paired speech inputs.
Table \ref{Tab: phone_align} shows that SpeechUT is able to align the most portion of units (about 85\%) with speech, where an interesting phenomenon shows that the alignment varies distinctly with respect to different kinds of phonemes.

\section{Conclusion}

In this paper, we propose SpeechUT, a unified-modal speech-unit-text pre-training model, which bridges the modality gap between speech  and text representation with hidden units.
By pre-training with the speech-to-unit task, masked unit modeling task, and unit-to-text task, SpeechUT significantly outperforms strong baselines as well as previous works and achieves state-of-the-art performance on downstream speech recognition and speech translation tasks.
In the future, we are interested in removing the dependence on a small amount of paired ASR data before pre-training, and extending SpeechUT to a multilingual model.

\section*{Limitations}
While the proposed SpeechUT model leverages hidden-unit representation as the bridge between speech and text, and obtains significant improvement over previous works, it still has some limitations: (1) the current method is a semi-supervised pre-training method, where the T2U generator needs paired ASR data to train, and takes external time to generate the units from the text; (2) the proposed SpeechUT only supports speech-to-text tasks, and it would be nice to able to help text-to-speech and speech-to-speech tasks; (3) we have to pre-train an independent model for each translation pair in the current method, which is time-consuming and resource-consuming; (4) the effectiveness of applying SpeechUT to other speech domains (e.g. child speech, accented speech) needs to be further investigated.

\section*{Ethics Statement}
This work presents SpeechUT, a pre-trained model for speech recognition and speech translation.
We evaluate our methods on standard benchmarks of the research community.
The datasets used in this study contain LibriSpeech, LibriLight, LibriSpeech LM Corpus, and MuST-C.
They are all public datasets and are widely used in the research community.

\bibliographystyle{acl_natbib}
\bibliography{anthology,emnlp2022_final}

\clearpage
\appendix
\section{Preliminary evaluation on hidden units} \label{Appen:cascade}
As a preliminary validation of using hidden units as an intermediate modality between speech and text, we 1) cascade a speech-to-unit model with a unit-to-text model for speech-to-text evaluation, and 2) cascade a text-to-unit model with a unit-to-text model for text-to-text evaluation.

The speech-to-unit model is a k-means model learned from the released HuBERT Base model, which is exactly the S2U generator introduced in Section \ref{ssec:unit-generator}.
The text-to-unit model is an encoder-decoder based sequence-to-sequence model trained on \textit{(speech-generated units, text)} paired data of LibriSpeech \texttt{train-clean-100}, which is exactly the T2U generator introduced in Section \ref{ssec:unit-generator}.
The unit-to-text model has the same architecture with the text-to-unit model but with reversed inputs/outputs.
The training data of the unit-to-text model comes from (1) \texttt{train-clean-100} subset like training text-to-unit model, and (2) pseudo unit-text data, in which we use text-to-unit model to generate pseudo unit from text data in LibriSpeech LM corpus.

The detailed results are listed in Table \ref{Tab:cascade}.
Although units lose some information (e.g., the repetitive frames are merged) of speech, it still achieves low WER (7.3/18.1) compared to the oracle speech-to-text (CTC) model.
On the other hand, cascading T2U and U2T models, which means translating text into units and then translating back, also achieves low WER.
These results indicate the units produced by the off-line S2U/T2U generators remain the main linguistic information of both speech and text, thus working as a bridge between the two modalities.

\begin{table}[h!]
\centering
\small
\begin{tabular}{c|cccc}
\toprule
\multirow{2}{*}{Model}  & \multicolumn{2}{c}{Dev}   & \multicolumn{2}{c}{Test}      \\
                        & clean     & other         & clean     & other             \\
\midrule
\multicolumn{5}{l}{\textit{Directly fine-tune CTC model from HuBERT}} \\
S2T                     & -            & -           & 6.3          & 13.2         \\
\midrule
\multicolumn{5}{l}{\textit{Cascade two off-line models}} \\
S2U $\rightarrow$ U2T   & 6.9          & 17.9        & 7.3          & 18.1         \\
T2U $\rightarrow$ U2T   & 5.1          & 3.1         & 4.5          & 5.5       \\
\bottomrule
\end{tabular}
\caption{\label{Tab:cascade} WER between the true text and the generated text by different models.}
\end{table}

\section{Data statistics} \label{Appen:data}
All the data used in our experiments are listed in Table \ref{Tab:dataset}.
For LibriSpeech LM data, the text is directly processed into characters and sent to the T2U generator.
For WMT data which is much noisier, we normalize the English-side text by removing punctuation and converting digits to spoken words before sending them to the T2U generator.
We only keep the samples shorter than 250 words.
When generating units from text, we filter out a few portions (about 15\%) of data by thresholding the token-averaged decoding likelihood.
The threshold is set to -0.666.

\begin{table*}[h]
\begin{center}
\resizebox{1.0\linewidth}{!}{
\begin{tabular}{cc|cc|cc|cc|cc|cc|cc}
\hline
\multicolumn{2}{c|}{\multirow{3}{*}{Task}} & \multicolumn{6}{c|}{Pre-training Data}                                                                & \multicolumn{2}{c|}{T2U training data}     & \multicolumn{4}{c}{Fine-tuning   Data}            \\
\cline{3-14}
\multicolumn{2}{c|}{}                      & \multicolumn{2}{c|}{Unlabeled Speech} & \multicolumn{2}{c|}{Unpaired Text} & \multicolumn{2}{c|}{MT} & \multicolumn{2}{c|}{ASR}                   & \multicolumn{2}{c|}{ASR} & \multicolumn{2}{c}{ST} \\
\multicolumn{2}{c|}{}                      & name             & \#hour             & name            & \#utt           & name       & \#utt     & name                & \#hour              & name      & \#hour      & name       & \#hour    \\
\hline
\multirow{2}{*}{ASR}        & Base         & LS               & 960                 & LS LM           & 40M              & -          & -          & LS                  & 100                  & LS        & 100          & -          & -         \\
                            & Large        & LL               & 60k                 & LS LM           & 40M              & -          & -          & LS                  & 100                  & LS        & 100/960      & -          & -         \\
\hline
\multirow{3}{*}{ST}         & En-De        & LS, MuST-C       & 1.4k                & -               & -                & WMT16      & 4.6M       & LS                  & 100                  & -         & -            & MuST-C     & 408       \\
                            & En-Es        & LS, MuST-C       & 1.5k                & -               & -                & WMT13      & 15.2M      & LS                  & 100                  & -         & -            & MuST-C     & 504       \\
                            & En-Fr        & LS, MuST-C       & 1.5k                & -               & -                & WMT14      & 40.8M      & LS                  & 100                  & -         & -            & MuST-C     & 492       \\
\hline
\end{tabular}
}
\end{center}
\caption{\label{Tab:dataset} Statistics of datasets used in experiments. LS: LibriSpeech, LL: LibriLight.}
\end{table*}

\begin{table*}[h]
\begin{center}
\resizebox{\linewidth}{!}{
\begin{tabular}{lc|ccc|cc|ccc}
\toprule
\multirow{2}{*}{Model}                 & \multirow{2}{*}{Size}  & \multicolumn{3}{c|}{Pre-training Data}        & \multicolumn{2}{c|}{WER ($\downarrow$) Without LM} & \multicolumn{3}{c}{WER ($\downarrow$) With LM}   \\
                                       &                        & Speech           & Paired         & Text      & test-clean        & test-other        & LM            & test-clean   & test-other \\
\midrule
wav2vec 2.0 \cite{baevski2020wav2vec}  & Large (0.3B)           & 60kh    & -               & -                 & 2.2               & 4.5               & Transf.       & 1.8        & 3.3  \\
HuBERT \cite{hsu2021hubert}            & Large (0.3B)           & 60kh    & -               & -                 & -                 & -                 & Transf.       & 1.9        & 3.3  \\
WavLM \cite{chen2021wavlm}             & Large (0.3B)           & 94kh    & -               & -                 & -                 & -                 & Transf.       & 1.8        & 3.2  \\
ILS-SSL \cite{Wang9747022ILS}          & Large (0.3B)           & 60kh    & -               & -                 & 1.9               & 3.8               & Transf.       & 1.8        & 3.2  \\
STPT \cite{tang2022unified}            & Base (0.16B)           & 60kh    & 960h            & 40M               & 2.1               & 4.6               & Unknown       & 2.1        & 4.5 \\
w2v-Conformer \cite{zhang2020pushing}  & X-Large (0.6B)         & 60kh    & -               & -                 & 1.7               & 3.5               & LSTM.         & 1.5        & 3.2  \\
w2v-Conformer \cite{zhang2020pushing}  & XX-Large (1.0B)        & 60kh    & -               & -                 & 1.6               & 3.3               & LSTM.         & 1.5        & 3.1  \\
w2v-BERT \cite{Chung9688253w2vBERT}    & X-Large (0.6B)         & 60kh    & -               & -                 & 1.5               & 2.9               & LSTM.         & 1.5        & 2.8  \\
w2v-BERT \cite{Chung9688253w2vBERT}    & XX-Large (1.0B)        & 60kh    & -               & -                 & 1.5               & 2.8               & LSTM.         & 1.5        & 2.7  \\
SLAM \cite{bapna2021slam}              & X-Large (0.6B)         & 60kh    & 960h            & mC4-En            & 1.6               & 3.1               & -             & -          & -    \\
Maestro \cite{chen2022maestro}         & X-Large (0.6B)         & 60kh    & $\sim$5kh       & 54M               & 1.5               & 2.8               & Conf.         & 1.5        & 2.7 \\
\midrule
\textbf{SpeechUT (Ours)}               & Large (0.38B)          & 60kh    & 100h$^\dagger$    & 40M               & 1.6               & 3.6               & Transf.       & 1.6        & 3.0 \\
\bottomrule
\end{tabular}
}
\end{center}
\caption{\label{Tab: wer_960h} ASR performance on 960-hour LibriSpeech benchmark. Speech/Paired/Text indicates the unlabeled speech data, the paired ASR data, and the unpaired text data respectively. $^\dagger$ indicates the data is not directly used for pre-training. Transf./LSTM./Conf. indicate the Transformer/LSTM/Conformer language models.}
\end{table*}

\section{Training details} \label{Appen:training}
\paragraph{Pre-training}
For the base model, the pre-training is conducted on 32 V100 GPUs with the update frequency of 1.
The max-tokens of S2U, U2T, and MUM tasks on each GPU are 1,400,000 (87.5 seconds), 3,000, and 3,000, respectively.
We use Adam optimizer.
The maximum learning rate is $5e-4$ and increases linearly in the first 32K steps, then decays linearly to zero in the total 400k steps.
All modules are randomly initialized before pre-training.
The pre-training takes about 3 days.

For the large model, the pre-training is conducted on 64 V100 GPUs with the update frequency of 2.
The max-tokens are set to 900,000 (56.25 seconds), 2000, and 2000 for S2U, U2T, and MUM tasks respectively.
Other optimization configurations are the same with the base model.
The pre-training takes about 12 days.

\paragraph{ASR fine-tuning}
Due to the limitation of the GPU memory, the max tokens are set to 1,300,000 (81.25 seconds).
During fine-tuning, the speech masking probability is set to 5\%.
We use the tri-stage learning-rate scheduler with (warm-up, hold, decay) periods of ($10\%, 40\%, 50\%$).
The maximum learning rate is set to $1e-5$.
The base model is fine-tuned on 8 GPUs with the update frequency of 2 for 40k steps.
The large model is fine-tuned on 8 GPUs with the update frequency of 3 for 80k steps.

\paragraph{ST fine-tuning}
The max-tokens are set to 800,000 (50 seconds) due to the GPU memory and we drop the training samples longer than it.
The speech masking probability is set to 5\%.
The label smoothing is set to 0.1.
The learning rate increases linearly to $3e-5$ in the first 5K steps, then decays linearly to zero in total 50k steps.
Models are fine-tuned on 8 GPUs with the update frequency of 4.


\section{Inference details} \label{Appen:inference}
\paragraph{ASR inference}
We select the model with the highest accuracy on \texttt{dev-other} set as the final model and apply the joint CTC/ED decoding \cite{hori-etal-2017-joint}.
We also use a character-level Transformer language model (LM) for shallow fusion \cite{gulcehre2015using}, which is provided by \citet{ao2021speecht5} \footnote{\url{https://github.com/microsoft/SpeechT5}}.
According to \citet{ao2021speecht5}, the LM has a similar or higher word-level perplexity (means worse) than the Transformer LM used in the previous works in Table \ref{Tab: wer_100h}, the latter is provided by \citet{synnaeve2019end}.
During decoding, the beam size is set to 30 with LM fusion and 10 without it.
The ED weight, CTC weight and the LM weight are set to (0.7, 0.3, 0.7) and (0.8, 0.2, 0) respectively after searching on \texttt{dev-other} set.
\paragraph{ST inference}
We average the parameters of the last 10 checkpoints for inference.
The decoding beam is 10.
We report the case-sensitive detokenized BLEU \cite{papineni2002bleu} on \texttt{tst-COMMON} set.

\section{ASR results on 960-hour dataset} \label{Appen:960}
Table \ref{Tab: wer_960h} lists the results of the SpeechUT Large model fine-tuned on full LibriSpeech 960 hours of ASR data, compared with several previous self-supervised methods.
SpeechUT Large outperforms the previous works in the large model setting like wav2vec 2.0 \cite{baevski2020wav2vec}, HuBERT \cite{hsu2021hubert}, WavLM \cite{chen2021wavlm}, and ILS-SSL \cite{Wang9747022ILS}.
While, w2v-Conformer \cite{zhang2020pushing}, w2v-BERT \cite{Chung9688253w2vBERT}, SLAM \cite{bapna2021slam}, and Maestro \cite{chen2022maestro} use much larger models with Conformer blocks and/or more speech/text data, which are beyond fair comparison.

\end{document}